\documentclass[sigconf,authorversion]{acmart}
\AtBeginDocument{%
  \providecommand\BibTeX{{%
    \normalfont B\kern-0.5em{\scshape i\kern-0.25em b}\kern-0.8em\TeX}}}

\copyrightyear{2023}
\acmYear{2023}
\setcopyright{acmlicensed}
\acmConference[GeoAI '23]{6th ACM SIGSPATIAL International Workshop on AI for Geographic Knowledge Discovery}{November 13, 2023}{Hamburg, Germany}
\acmBooktitle{6th ACM SIGSPATIAL International Workshop on AI for Geographic Knowledge Discovery (GeoAI '23), November 13, 2023, Hamburg, Germany}
\acmPrice{15.00}
\acmDOI{10.1145/3615886.3627740}
\acmISBN{979-8-4007-0348-5/23/11}

\usepackage{listings}
\usepackage{graphicx}
\usepackage{booktabs}

\definecolor{codegreen}{rgb}{0,0.6,0}
\definecolor{codegray}{rgb}{0.5,0.5,0.5}
\definecolor{codepurple}{rgb}{0.58,0,0.82}
\definecolor{backcolour}{rgb}{0.95,0.95,0.92}

\lstdefinestyle{mystyle}{
    backgroundcolor=\color{backcolour},   
    commentstyle=\color{codegreen},
    keywordstyle=\color{magenta},
    numberstyle=\tiny\color{codegray},
    stringstyle=\color{codepurple},
    basicstyle=\ttfamily\footnotesize,
    breakatwhitespace=false,         
    breaklines=true,                 
    captionpos=b,                    
    keepspaces=true,                 
    numbers=none,                    
    numbersep=5pt,                  
    showspaces=false,                
    showstringspaces=false,
    showtabs=false,                  
    tabsize=2
}

\begin{document}

\title{SRAI: Towards Standardization of Geospatial AI}

\author{Piotr Gramacki}
\orcid{0000-0002-4587-5586}
\authornote{Equal contribution, authors in alphabetic order.}

\author{Kacper Leśniara}
\orcid{0000-0003-0875-4301}
\authornotemark[1]

\author{Kamil Raczycki}
\orcid{0000-0002-3715-4869}
\authornotemark[1]

\author{Szymon Woźniak}
\orcid{0000-0002-2047-1649}
\authornotemark[1]
\affiliation{%
  \institution{Wrocław University of Science and Technology}
  \department{Department of Artificial Intelligence / Kraina.AI}
  \city{Wrocław}
  \country{Poland}
}
\email{piotr.gramacki@pwr.edu.pl}
\email{{klesniara,kraczycki,swozniak}@kraina.ai}

\author{Marcin Przymus}
\orcid{0009-0004-7741-8541}
\author{Piotr Szymański}
\orcid{0000-0002-7733-3239}
\affiliation{%
  \institution{Wrocław University of Science and Technology}
  \department{Department of Artificial Intelligence / Kraina.AI}
  \city{Wrocław}
  \country{Poland}
}
\email{mprzymus@kraina.ai}
\email{piotr.szymanski@pwr.edu.pl}


\renewcommand{\shortauthors}{Gramacki, Leśniara, Raczycki and Woźniak, et al.}

\begin{abstract}
    Spatial Representations for Artificial Intelligence (\textit{srai}) is a Python library for working with geospatial data. The library can download geospatial data, split a given area into micro-regions using multiple algorithms and train an embedding model using various architectures. It includes baseline models as well as more complex methods from published works. Those capabilities make it possible to use \textit{srai} in a complete pipeline for geospatial task solving. The proposed library is the first step to standardize the geospatial AI domain toolset. It is fully open-source and published under Apache 2.0 licence.
\end{abstract}



\begin{teaserfigure}
  \includegraphics[width=\textwidth]{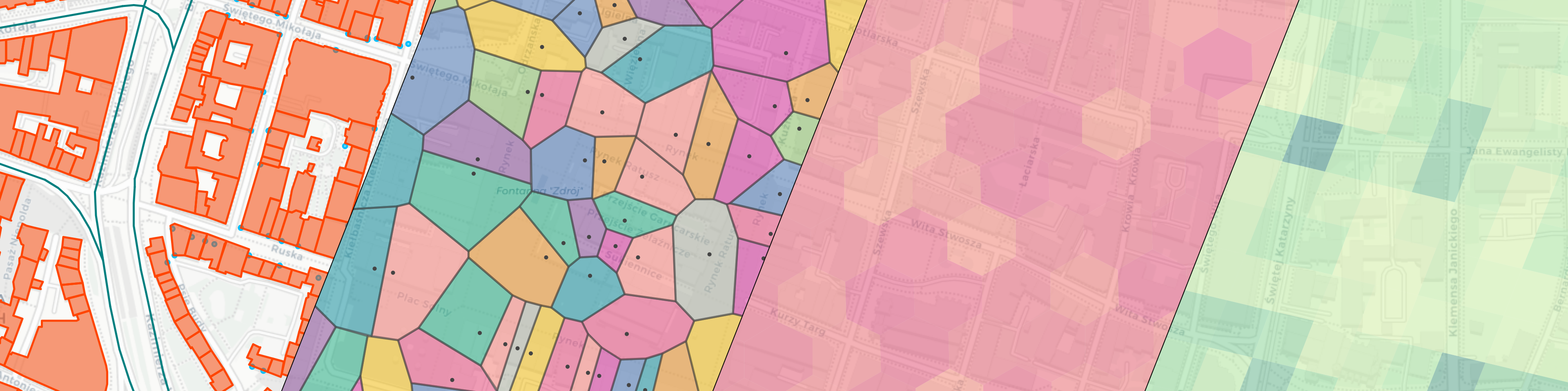}
  \caption{The \textit{srai} library features overview: data acquisition, area regionalization, and shape agnostic embedding methods.}
  \Description{A part of Wrocław on which we present four ways of space division - building boundaries, Voronoi regions anchored around restaurant locations, H3 and S2 cells.}
  \label{fig:teaser}
\end{teaserfigure}

\maketitle

\section{Introduction}

Spatial Representations for Artificial Intelligence (\textit{srai}) is a fully open-source Python module integrating many geo-related algorithms in a single package with a unified API. It aims to provide simple and efficient solutions to geospatial problems - accessible to everybody and reusable in various contexts. Many tasks can benefit from enrichment with geospatial data. Our main goal is to unify geospatial data processing and models, making reproducing experiments and sharing new models easier. We have already included previously published models for geospatial representation learning. The library is available on GitHub\footnote{\url{https://github.com/kraina-ai/srai}} where one can find a link to pre-trained models download.  

With \textit{srai}, we want to lay the groundwork for unifying geospatial models for representation learning and make it possible to reuse them easily for different tasks. We hope that more researchers follow this trend and that the \textit{srai} library will take the first steps to standardize the geospatial AI domain. 

\subsection{Geospatial data and tasks}

The geospatial domain has recently gained popularity in academic research and business applications. The vast amount of data available becomes hard to analyze using traditional GIS methods.
Companies try to leverage internal data and combine them with external sources to gain more insight for location intelligence tasks such as site selection, market analysis, geo-marketing, supply chain optimization, fraud detection, real-time traffic management, and navigation\footnote{https://carto.com/solutions}. Governments and academic researchers try to improve safety and quality of life by modelling the risk of natural disasters\cite{intro:applications:natural-disasters}, analyzing air pollution\cite{intro:applications:air-quality, intro:applications:air-quality-2}, predicting forest fires\cite{intro:applications:forest-fires}, analysis of public transport networks\cite{intro:applications:public-transport-usage} or traffic\cite{intro:applications:traffic-prediction} and many more.

To be used in a geospatial context, data must contain information that either represents a geometry with coordinates or can be geocoded to a region or a location. That said, a lot of data available can be included in geospatial tasks. Those can take the form of text (tags, reviews, social media posts, documents), images (satellite multispectral and hyperspectral imagery, aerial photos, street view images, and videos), and numeric values (weather readings, number of residents in a household, average time spent in traffic, POI ratings).

The \textit{srai} library focuses mainly on publicly available datasets with clearly defined vector geometry (points, lines, polygons) to increase reproducibility, simplify usage and democratize the geospatial domain for everyone without the need to be an expert in the GIS tools usage.

\subsection{OpenStreetMap}

Regarding geospatial data, one of the most critical open data sources is OpenStreetMap\cite{}. It is a free, editable map service that allows people worldwide to post updates and improve data coverage. It contains a wide range of tags, which allows for selecting a subset suitable for a given task. We chose it as the primary data source available in \textit{srai} and provided multiple ways of accessing it.

\subsection{Representation learning}

We designed the \textit{srai} library around representation learning approaches to geospatial data. Representation learning is finding a function $f: D \rightarrow \rm{I\!R}^{d}$, which transforms data $D$ into a $d$-dimensional vector. Similar to Natural Language Processing (NLP)\cite{word2vec, elmo} and Computer Vision (CV)\cite{related:representation-learning-CV-review, related:representation-learning-on-CV-1}, this approach has already proven helpful in solving geospatial tasks\cite{related:spatial-embeddings:cape, related:spatial-embeddings:hrnr, related:spatial-embeddings:ze-mob}. A significant benefit of such representations is that they are reusable. They can either be transferred from task to task or defined as general enough to be used in multiple further downstream tasks (similarly to pre-trained transformer models in NLP).

The remainder of this paper is organized as follows: Section\,2 presents an overview of available solutions in the geospatial area and positions the \textit{srai} library in comparison to them; Section 3 describes the library and explains the design choices we made; Section 4 presents the use cases for \textit{srai}, to show how it can improve geospatial tasks solving; Section 5 discusses future improvements in \textit{srai} to show the vision for the library; Section 6 concludes this paper and discusses the limitations of \textit{srai} in its current state.

\section{Background}

This section provides an overview of existing methods for geospatial analysis. It consists of tools we integrated into \textit{srai} and other libraries that simplify geospatial data processing. We also show how \textit{srai} compares to those alternatives.

This review is composed of two parts: (i) a description of geospatial tools that power \textit{srai} library and (ii) positioning of \textit{srai} relative to geospatial tools and systems of different types. 

\subsection{Ecosystem behind srai}

In this part, we describe the tools (both geospatial and not) we used to make \textit{srai} versatile and capable. We divided them into three categories: (i) \textit{Geospatial operations}, (ii) \textit{Data acquisition and processing}, and (iii) \textit{Machine Learning tooling}. We describe each of them in detail in the subsequent sections. 

\subsubsection{Geospatial operations}

The main workhorse in \textit{srai} is the GeoPandas\cite{tools:geopandas} library. It is a library compatible with pandas, which extends it with geospatial operations. The ease of integration with other tools and versatile capabilities was why we selected this library. Geometries are defined using Shapely\cite{tools:shapely}, which is the most popular geometry-related library in the Python domain and GeoPandas relies on it heavily. For regionalization (also referred to as tessellation in the literature), we use either our own implementations or pre-defined hierarchical spatial indexes - H3\cite{h3} and S2\cite{s2}. Both of them are widely utilized in the community, especially H3, which is why we included them.

\subsubsection{Data acquisition and processing}

The second group of tasks handled using external libraries in \textit{srai} is downloading and processing spatial data. The primary data source in \textit{srai} is OpenStreetMap, which we access using the OSMnx\cite{tools:osmnx} library. For loading OSM data from PBF files, we use pyosmium\cite{tools:pyosmium} and DuckDB\cite{tools:duckdb}. Reading a vast majority of geospatial file formats is enabled by GeoPandas.

Another data source available in \textit{srai} is GTFS timetables, which we process using the GTFS Kit\cite{tools:gtfs-kit} library. It provides a wide range of operations, which lifts operating on relational GTFS schemes off our shoulders and provides calculations for basic public transport statistics. One of its more convenient capabilities is GTFS feed validation, which we also use in \textit{srai}.

\subsubsection{Machine Learning tooling}

Apart from geospatial operations, we also use multiple libraries for data science and machine learning. We use widely adopted libraries to make \textit{srai} compatible with the most popular tools and frameworks. Also, leveraging well-known libraries makes our library easier for first-time users. Our operations are backed by: pandas\cite{tools:pandas} and NumPy\cite{tools:numpy} for data processing and calculations; scikit-learn\cite{tools:scikit-learn} for machine learning algorithms; NetworkX\cite{tools:networkx} for graph processing; PyTorch\cite{tools:pytorch} with PyTorch Lightning\cite{tools:pytorch-lightning} for deep learning models.

\begin{table*}[t]
\caption{Geospatial toolboxes and their capabilities.}
\label{tab:tools:toolboxes}
\resizebox{\textwidth}{!}{\begin{tabular}{@{}l|ccccc|ccc|cc@{}}
\toprule
Library                                                 & Spatial files        & OSM        & Trajectories & GTFS         & Raster         & Visualization     & Regionalization   & Geocoding  & ML         & Datasets   \\ \midrule
geowrangler$^1$                                         & \checkmark           & \checkmark &              &              & \checkmark     &                   & \checkmark        &            & \checkmark &            \\
tesspy$^2$                                              &                      & \checkmark &              &              &                & \checkmark        & \checkmark        & \checkmark &            &            \\
geomancer$^3$                                           &                      & \checkmark &              &              &                &                   &                   &            &            &            \\
Mosaic$^4$                                              & \checkmark           &            &              &              & \checkmark     &                   & \checkmark        &            & \checkmark &            \\
PySal\cite{toolboxes:pysal}                             &                      &            &              &              & \checkmark     & \checkmark        & \checkmark        &            & \checkmark &            \\
Verde\cite{toolboxes:verde}                             & \checkmark           &            &              &              &                &                   & \checkmark        &            & \checkmark & \checkmark \\
WhiteboxTools$^6$                                       & \checkmark           &            &              &              & \checkmark     &                   &                   &            & \checkmark &            \\
Pandana$^5$                                             & \checkmark           & \checkmark &              &              &                &                   &                   &            &            &            \\
MovingPandas\cite{toolboxes:moving-pandas}              &                      &            & \checkmark   &              &                &                   &                   &            &            &            \\
Scikit mobility\cite{toolboxes:scikit-mobility}         &                      &            & \checkmark   &              &                &                   &                   &            &            &            \\
segment-geospatial\cite{toolboxes:segment-geospatial}   & \checkmark           &            &              &              & \checkmark     &                   &                   &            & \checkmark &            \\
TorchGeo\cite{tools:torchgeo}                           & \checkmark           &            &              &              & \checkmark     &                   &                   &            & \checkmark & \checkmark \\ \midrule
srai                                                    & \checkmark           & \checkmark &              & \checkmark   & \checkmark*     & \checkmark        & \checkmark        & \checkmark & \checkmark &            \\ \bottomrule
\end{tabular}}
\caption*{\tiny{
$^1$ \url{https://github.com/thinkingmachines/geowrangler}, 
$^2$ \url{https://github.com/siavash-saki/tesspy},
$^3$ \url{https://github.com/thinkingmachines/geomancer},
$^4$ \url{https://github.com/databrickslabs/mosaic},
$^5$ \url{https://github.com/UDST/pandana},
$^6$ \url{https://github.com/jblindsay/whitebox-tools},
* only for data downloading and preparation
}}
\end{table*}

\subsection{Positioning of srai}

Let us juxtapose \textit{srai} with existing tools for geospatial applications. We have identified a wide selection of libraries and systems in the geospatial domain and categorized them into four categories: (i) \textit{GIS Platforms}, (ii) \textit{Location Intelligence Platforms}, (iii) \textit{Spatial Databases and Computing Engines} and (iv) \textit{Toolboxes}.

\textit{GIS Platforms} are great, versatile tools for geospatial processing. Those are complete solutions that cover many use cases in the domain. The most popular and widely adopted, according to our review, are: ArcGIS\footnote{\url{https://www.arcgis.com/index.html}}, QGIS\footnote{\url{https://www.qgis.org/}} and Google Earth Engine\footnote{\url{https://earthengine.google.com/}}. Those systems have a different scope than the \textit{srai} library - ArcGIS and QGIS tend to be an all-purpose solution to geospatial data, while Google Earth Engine's focus is on handling raster data and providing access to a variety of geospatial datasets. These tools offer APIs in either Python or JavaScript or use them as scripting languages inside the tools themselves. Regarding pricing: QGIS is free and open-source, Google Earth Engine is free for noncommercial use, and ArcGIS is a paid solution. Compared to fully-fledged GIS applications, the \textit{srai} library is free and open-source. It depicts itself more as a specific set of functionalities targeted at the geospatial ML domain and simplifying data downloading and wrangling. The advantage is that it is accessible for people without a GIS background, such as data scientists and Python developers.

\textit{Location Intelligence Platforms} is a set of systems that collect data and enable analytics, mainly in the Location Intelligence domain. 
They prioritize cloud-based analytics, spatial data enrichment, and online visualizations.
Examples are CARTO\footnote{\url{https://carto.com/}} and Foursquare\footnote{\url{https://foursquare.com/}}, which focus more on location intelligence, as well as Mapbox\footnote{\url{https://www.mapbox.com/}} which is primarily a map and navigation provider. All solutions provide a data catalogue with geo-related datasets from domains such as traffic, movement, demographics, weather, retail, etc. Additionally, those platforms expose APIs for creating their map-based applications. Access to these platforms requires a paid subscription.
Those platforms are very popular in the geospatial domain, especially in the corporate world, but need more machine learning functionalities.
The \textit{srai} library allows for creating more sophisticated machine learning models. Additionally, using paid, quality datasets from those platforms could create more advanced solutions in business applications and extend the capabilities of those platforms.

\textit{Spatial Databases and Computing Engines} are dedicated solutions for large-scale geospatial computations and data storage. Examples include PostGIS\footnote{\url{http://postgis.net/}} which is a spatial extension to PostgreSQL\footnote{\url{https://www.postgresql.org/}} and Apache Sedona\footnote{\url{https://sedona.apache.org/latest-snapshot/}} which is a highly efficient solution for spatial data based on Spark\footnote{\url{https://spark.apache.org/}}. There are also cloud-based data warehouses capable of processing geospatial data, such as BigQuery\footnote{https://cloud.google.com/bigquery}, Athena\footnote{https://aws.amazon.com/athena/} and Snowflake\footnote{\url{https://www.snowflake.com/en/}}, which simplify set-up and maintenance.
Alternatively, large-scale datasets can be handled using either parallelism-capable libraries (e.g., Dask-GeoPandas\footnote{\url{https://github.com/geopandas/dask-geopandas}}) or more performant ones (e.g., GeoPolars\footnote{\url{https://github.com/geopolars/geopolars}})
All of the mentioned solutions could be treated as possible backends for the \textit{srai} library regarding storage and/or computation, as it is with GeoPandas and DuckDB.

\textit{Toolboxes} is a coherent collection of functionalities dedicated to a limited set of use cases. We have gathered a wide selection of such tools, compared in Table\,\ref{tab:tools:toolboxes}. Most of the depicted tools individually contain a subset of the \textit{srai} functionalities. However, none offers complete coverage of the geospatial AI pipeline. 
The library most resembling \textit{srai} is wrangler. It offers similar data loading and processing capabilities but is less ML-oriented and offers fewer utility functions such as plotting or geocoding. Mosaic also overlaps with \textit{srai}. However, it is only available on the Databricks platform. We have identified areas that are not covered in \textit{srai} - trajectories (MovingPandas and scikit-mobility) and raster-focused data processing and machine learning (TorchGeo and segment-geospatial). We have also found libraries focused primarily on feature engineering and analytics - geomancer, PySal, Pandana, WhiteboxTools - or pure regionalization - tesspy. In comparison to the toolboxes mentioned above, \textit{srai} introduces a needed enhancement to the geospatial tooling available to data scientists and developers.

\section{The srai library}

This section explains the design of the library and lists methods which are already available in \textit{srai}.

\subsection{General library design}

The \textit{srai} library design aims to bring different geospatial operations under a common interface. With that achieved, different algorithms will be easily interchangeable in the geospatial task pipeline. We define four main components:

\begin{enumerate}
    \item Loader - responsible for loading geospatial data (features) from a given source and pre-processing it to a format used in \textit{srai},
    \item Regionalizer - able to split a given area into micro-regions used in further analysis,
    \item Joiner - used to match loaded features with regions based on a given spatial predicate,
    \item Embedder - which embeds regions into a vector space based on features matched to them.
\end{enumerate}

\begin{figure}
    \centering
    \includegraphics[width=0.45\textwidth]{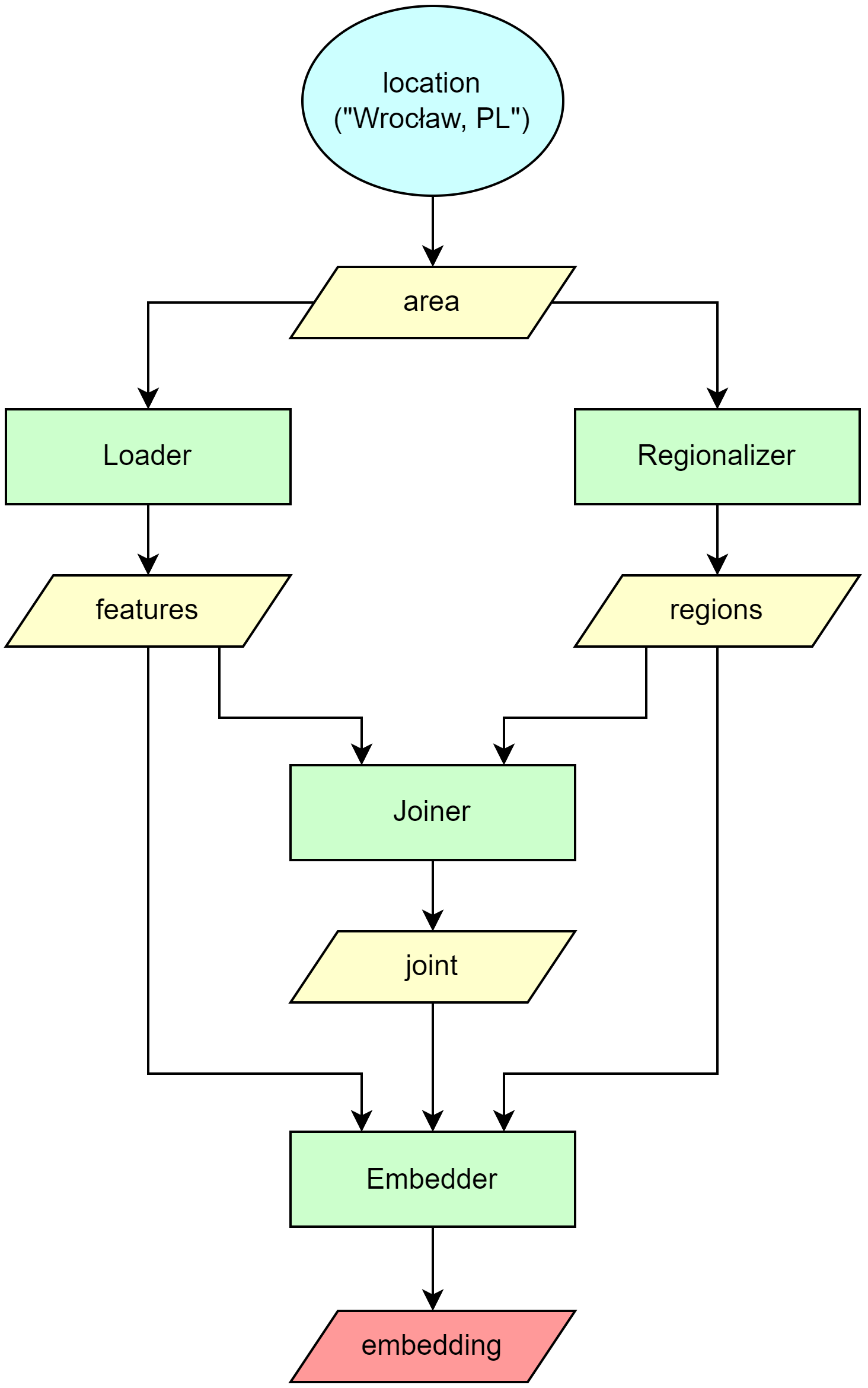}
    \caption{Embedding calculations for a given location using the \textit{srai} library.}
    \label{fig:lib-diagram}
\end{figure}

Those components create a complete pipeline for learning representations for geospatial data. Solving the final task can be obtained using any machine learning or deep learning library since \textit{srai} returns embeddings as pandas\cite{tools:pandas} data frames, which are very versatile. The embedding step can be omitted, using a dedicated component that only summarises each region's features. A general overview of how those components are connected is presented in Figure\,\ref{fig:lib-diagram}. We start with location definition, which we can geocode using our utility functions. Next, we use a \textit{Loader} to get geospatial features (for example, from OSM) from the obtained area and a \textit{Regionalizer} to split this area into micro-regions (for instance, H3 cells). Regions are combined with features using a \textit{Joiner}, and the results are passed over to an \textit{Embedder}, which produces the final embeddings.

\subsection{Functionalities in srai}

The \textit{srai} library implements all steps of the representation learning approach based on OSM data. In almost all steps, there are multiple options implemented, which are interchangeable thanks to unified interfaces. 

\subsubsection{Loaders}

The first step is to gather data. In \textit{srai}, our main source of data is OpenStreetMap, and we provide tools to download different data from there:

\begin{itemize}
    \item OSM tags - downloading tags specified by a filter for a given area. We implement queries to OSM online (suitable for a small number of tags) and from \textit{PBF} files (more efficient in cases with more tags). We provided a convenient functionality for users that automatically selects the smallest extract with OSM data to download, based on the given geometries, and downloads it either from \mbox{Geofabrik\footnote{https://www.geofabrik.de/}} or OpenStreetMap\cite{OpenStreetMap} hosting services.
    \item OSM networks - downloading structured networks as a graph. It can download different types of networks (roads, bike paths, etc.),
    \item OSM tiles - downloading tiles with maps as images. It works with different zoom levels and tile providers,
    \item GTFS - loading data from GTFS feeds and pre-computations of public transport offer for stops (based on gtfs2vec\cite{our:gtfs2vec} paper),
    \item GeoParquet - loading geospatial data from \textit{geoparquet} files, with pre-processing required by other library components.
\end{itemize}

\subsubsection{Regionalizers}

\begin{figure}
    \centering
    \includegraphics[width=0.49\textwidth]{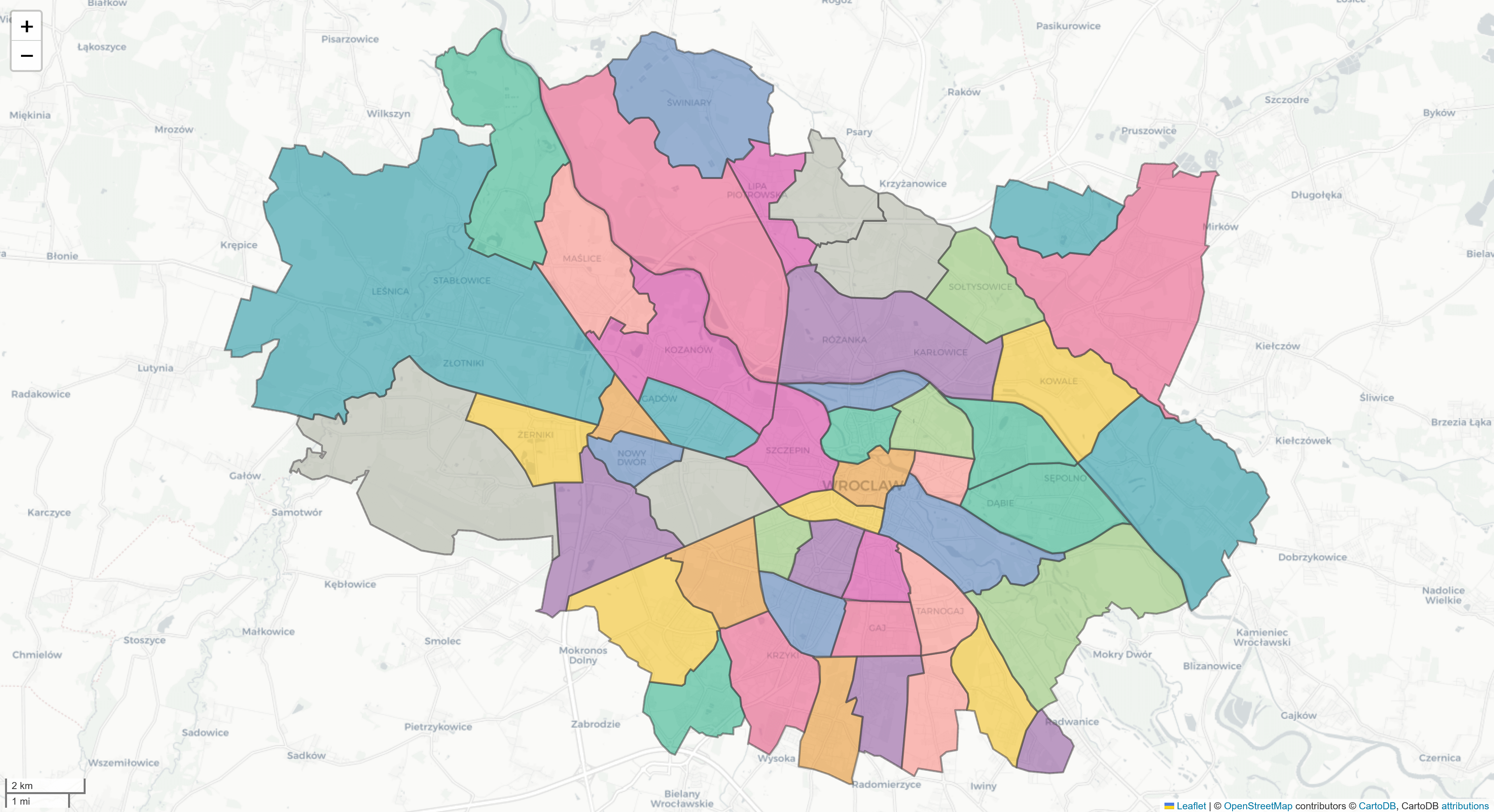}
    \caption{Wrocław city districts downloaded by Administrative Boundary Regionalizer.}
    \label{fig:administative-regions}
\end{figure}

The next step is to split the area of interest into micro-regions, for which the data will be aggregated, and representations will be computed. For this task, we implement support for spatial indexes as well as data-based division approaches:

\begin{itemize}
    \item H3 and S2 - spatial indices with different shapes, hexagons, and squares, respectively. Both are deterministic and implemented hierarchically.
    \item Voronoi - allows the given areas to be divided into Thiessen polygons using seed geometries,
    \item Administrative boundaries - splits an area based on administrative boundaries on different levels available in OSM (see example in Figure \ref{fig:administative-regions}),
    \item Slippy map - splits an area into regions, which match with OSM tiles from the corresponding loader.
\end{itemize}

\subsubsection{Joiners}

Features returned by the loader should be matched with regions if regionalization is used in the pipeline. This can be done, with different spatial predicates; in \textit{srai}, we implement one which covers most of the use-cases - intersection.

\subsubsection{Embedders}

The key functionality of \textit{srai} is supporting geospatial data embedding. We provide multiple embedders:

\begin{itemize}
    \item Count - baseline embedder, which simply counts feature occurrences in regions,
    \item Contextual Count - counts features in regions with added information from neighbour regions (used by \cite{our:transfer-learning-bicycles}),
    \item Hex2Vec - embedder from hex2vec\cite{our:hex2vec} paper, which works with OSM tags,
    \item Highway2Vec - implementation from highway2vec\cite{our:highway2vec} paper of an embedder for road segments,
    \item GTFS2Vec - public transport offer embedding, based on the method described in the gtfs2vec\cite{our:gtfs2vec} paper.
\end{itemize}

\subsubsection{Pre-trained models}

Some of the embedders mentioned above require model training. This process can be resource-consuming, making the experiment challenging to reproduce. To overcome this problem, we include an option to save a pre-trained embedder to a file and quickly load it. We also share a selection of pre-trained hex2vec models and intend to extend this list in time. 

\subsubsection{Neighbourhoods}

Some of the geospatial tasks can benefit from including neighbourhood information. At the moment, two different neighbourhood implementations are available in \textit{srai}:

\begin{itemize}
    \item Adjacency - based on regions touching each other
    \item H3 - based on the H3 spatial index
\end{itemize}

\subsubsection{Utilities}

Apart from those main components, \textit{srai} contains various utility functions, which speed up spatial data processing or unify data formats which are used. This includes geocoding, buffering, flattening, merging, plotting, and other operations.

\begin{figure}
    \centering
    \includegraphics[width=0.49\textwidth]{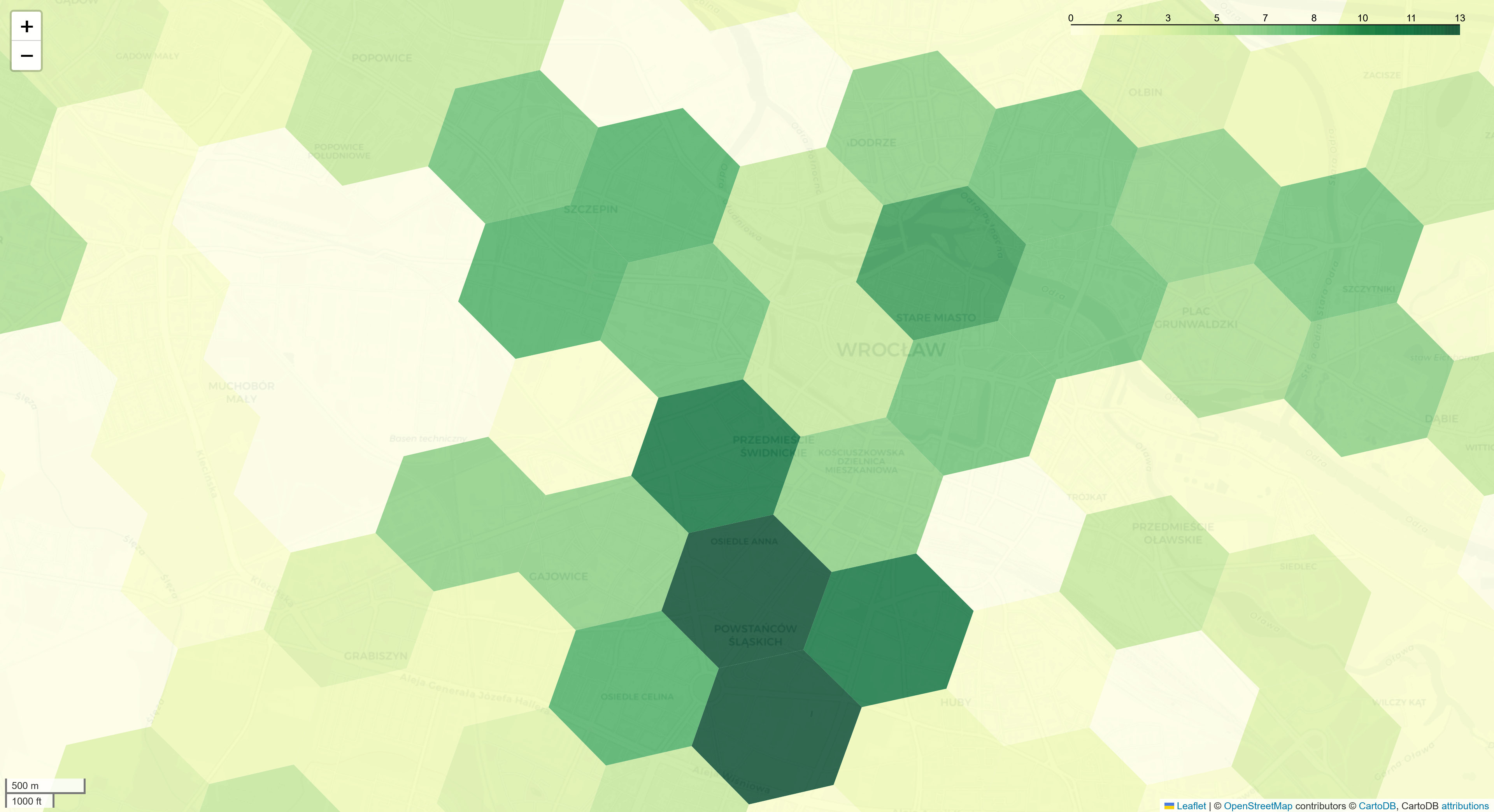}
    \caption{Sum of parks objects per H3 cell in Wrocław, Poland.}
    \label{fig:sum-of-parks}
\end{figure}

\subsection{Technical backend}

We need a spatial operations backend to make all of the abovementioned operations possible in \textit{srai}. We decided to use GeoPandas as our primary spatial backend. It is efficient for smaller use cases, which can be fitted into the available memory. One of the most important benefits is its compatibility with the pandas interface, which makes it easy to integrate with existing libraries. This high compatibility is crucial to making \textit{srai} a standard geospatial AI library. However, this approach is limited by available memory since all data must be loaded into RAM. This limitation makes it impossible to process larger areas simultaneously, so we are actively working on alternative geospatial backends. We have already started working on utilization of DuckDB's spatial extension for most memory-consuming tasks. We provide more detail on that in Section \ref{future:duckdb}.

\section{Usage examples}

This section presents examples of \textit{srai} usage in various supported tasks. We show how the unification of interfaces makes solving geospatial tasks easier. We present examples for (i) geospatial data download and processing, (ii) unsupervised and (iii) supervised task solving, and (iv) \textit{srai} usage for transfer learning. We present \textit{srai} functionalities on a single task of rental price estimation. We assume that we have data from multiple cities with short-term rental offerings, containing location and price per person. We split the task into parts to present how to apply \textit{srai} to each of them.

\subsection{Processing pipeline}
\label{sec:examples:processing}
Firstly, we prepare both the input (features from OSM) and output (rental price) data. In this example, we try to predict the average price in an area defined as a single H3 cell based on leisure and amenities offered. Listing \ref{lst:processing} presents how to do this using our library. The steps are as follows:

\begin{enumerate}
	\item download the boundaries of the analyzed city,
	\item define OSM tags filter and download the features from OSM,
	\item split an area into H3 cells,
	\item aggregate OSM features for cells, creating pre-training data.
\end{enumerate}

In Figure \ref{fig:sum-of-parks} we can see an example result for these steps - the number of park objects in each of the hexagonal micro-regions we selected.

\begin{lstlisting}[language=Python, caption=Loading and processing data from OSM for a given area using srai library., label=lst:processing]
import geopandas as gpd
from srai.loaders import OSMOnlineLoader
from srai.regionalizers import H3Regionalizer
from srai.joiners import IntersectionJoiner
from srai.utils import geocode_to_region_gdf

# download boundaries of the city
area = geocode_to_region_gdf("Wroclaw, PL")

# define which tags to download
query = {
    "leisure": ["park"], 
    "amenity": ["restaurant"]
}

# use loader to download OSM data
loader = OSMOnlineLoader()
features = loader.load(area, query)

# split an area into H3 cells
regionalizer = H3Regionalizer(resolution=8)
regions = regionalizer.transform(area)

# match features to regions
joiner = IntersectionJoiner()
joint = joiner.transform(regions, features)

\end{lstlisting}

\subsection{Clustering}

Secondly, we perform an exploratory analysis. This helps us understand how our data is distributed in the city and presents \textit{srai} capabilities in solving unsupervised tasks. We use the hex2vec method and clustering for that. All of these steps are illustrated in Listing \ref{lst:clustering}:

\begin{enumerate}
	\item create and configure hex2vec model,
	\item train it on OSM features downloaded earlier,
	\item use KMeans clustering to find similar areas.
\end{enumerate}

\begin{figure}
    \centering
    \includegraphics[width=0.49\textwidth]{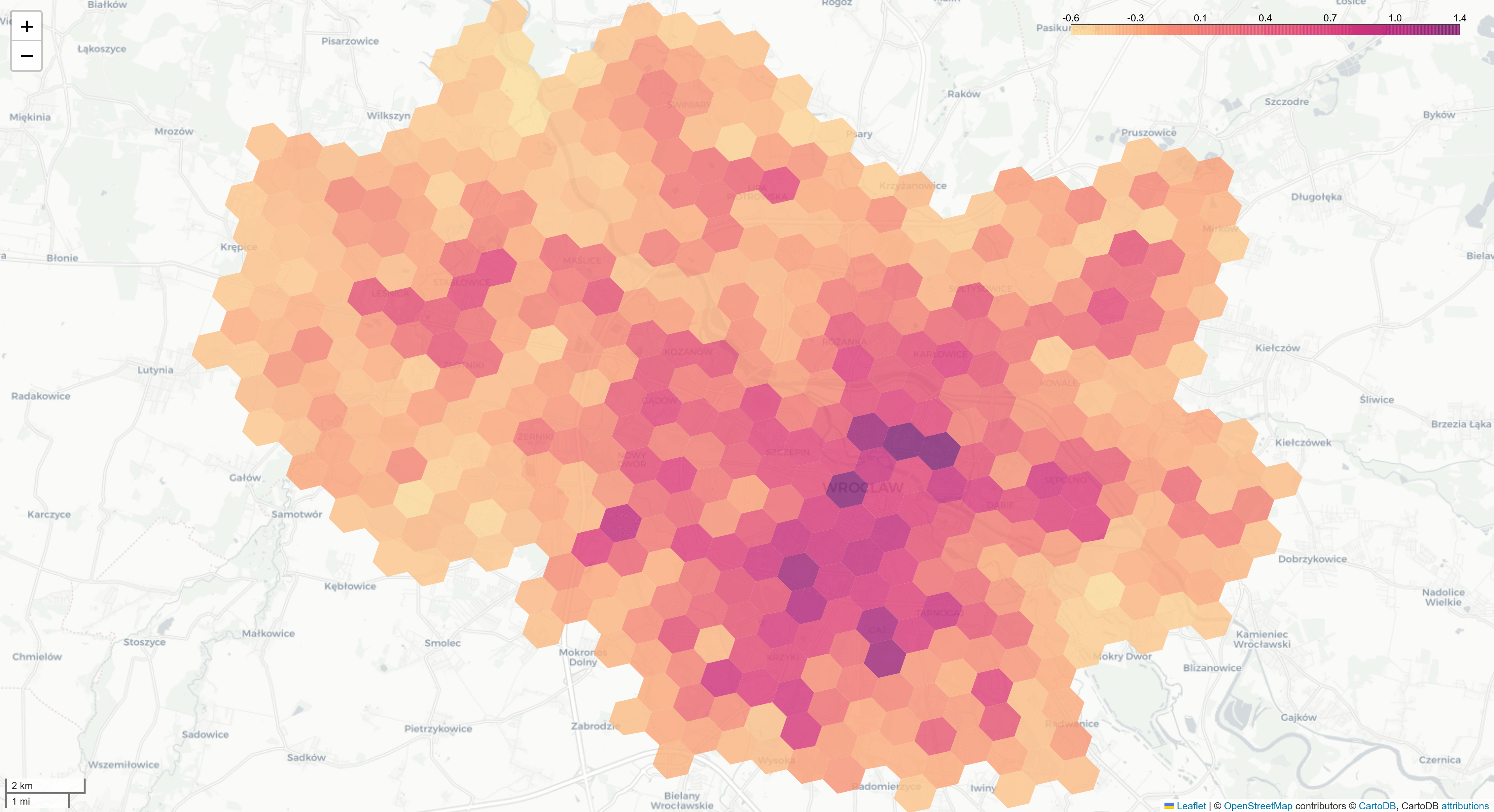}
    \caption{Hex2Vec model embedding layer visualization.}
    \label{fig:embedding-visualized}
\end{figure}

\begin{figure}[t]
    \centering
    \includegraphics[width=0.49\textwidth]{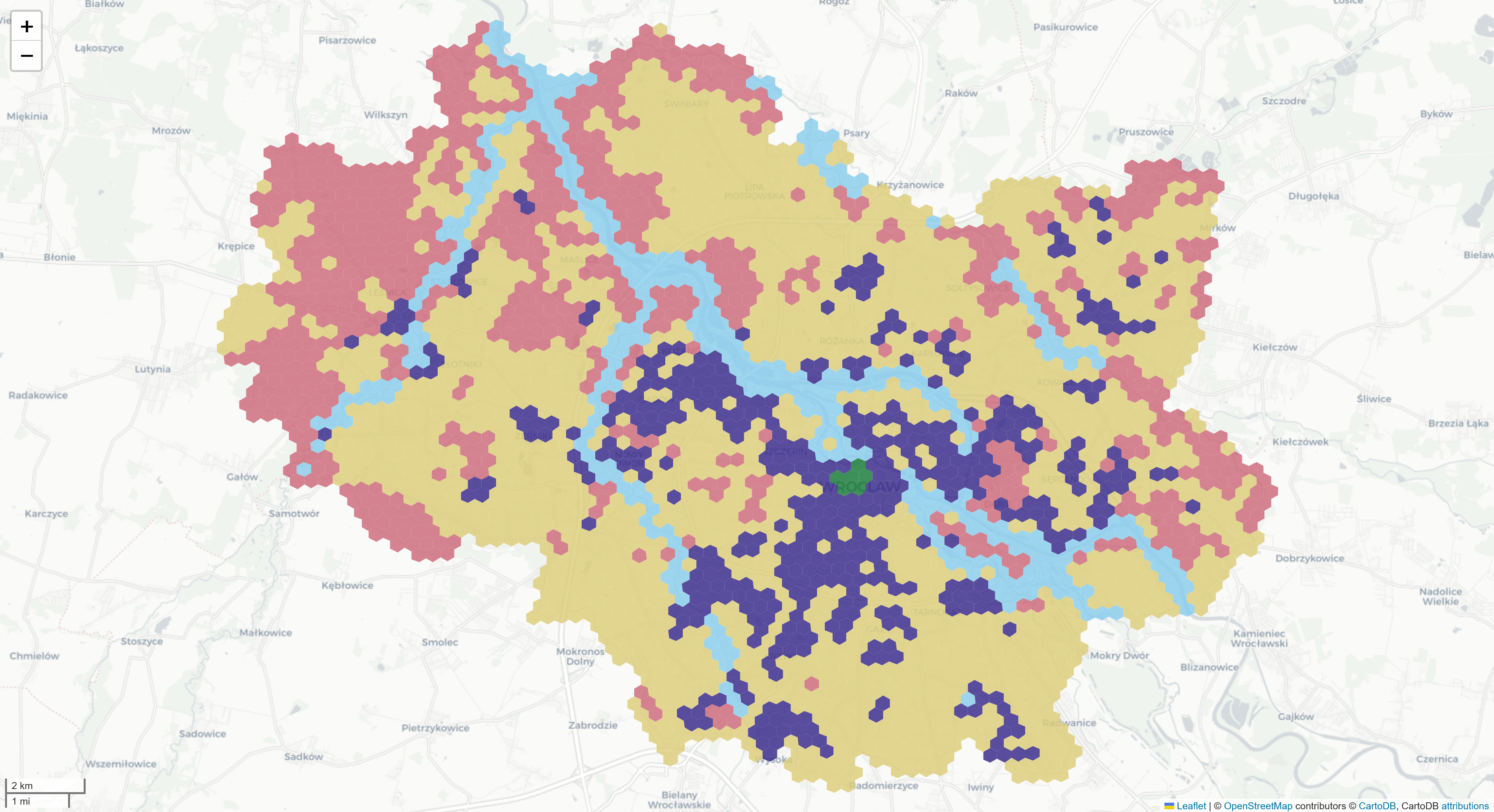}
    \caption{Clustering based on embeddings from \textit{srai} library.}
    \label{fig:clustering-result}
\end{figure}

Based on that result, we can make some preliminary observations on how rental price is affected by different leisure and amenities. Such an approach can also be useful in exploratory analysis and embedding method evaluation. In Figure \ref{fig:embedding-visualized} we can see a visualization of our model's embedding space and Figure \ref{fig:clustering-result} presents clustering results. Both of these visualizations can be useful in exploratory analysis before predictive model training.

\begin{lstlisting}[language=Python, caption=Training a hex2vec embedder and clustering in obtained embedding space, label=lst:clustering]
from srai.embedders import Hex2VecEmbedder
from srai.neighbourhoods import H3Neighbourhood
# use any library for clustering you want
from sklearn.cluster import KMeans

# configure hex2vec embedder
embedder = Hex2VecEmbedder(encoder_sizes=[42, 13])
neighbourhood = H3Neighbourhood(regions)

# pre-train hex2vec model on OSM data
embeddings = embedder.fit_transform(
    regions,
    features,
    joint,
    neighbourhood,
    batch_size=64,
)

# run clustering to explore differences between regions
clustering = KMeans(5)
clustering.fit(embeddings)
regions["cluster"] = clustering.labels_
\end{lstlisting}

\subsection{Prediction}

Next, we train a regression model to estimate rental prices. We use our pre-trained hex2vec model from the previous step and combine it with the regression head. Then we train these models on rental prices, which we preprocessed using our library. This is presented in Listing \ref{lst:prediction} and consists of the following steps:

\begin{enumerate}
    \item load prices dataset and preprocess it using \textit{srai},
    \item generate train and test data,
    \item create Regressor based on hex2vec embeddings,
    \item train the model and evaluate it.
\end{enumerate}

\begin{lstlisting}[language=Python, caption=Using pre-trained hex2vec embeddings to train a regressor for price estimation, label=lst:prediction]
import geopandas as gpd
from srai.constants import FEATURES_INDEX, REGIONS_INDEX
# use the library and regressor of your choice
from sklearn.svm import SVR
import sklearn.metrics.mean_squared_error as mse

# load rental prices and calculate avg per region
prices = gpd.read_file("prices.geojson")
prices.index.name = FEATURES_INDEX
pj = joiner.transform(regions, prices)
pj = pj.join(on=FEATURES_INDEX).reset_index(inplace=True)
prices_avg = pj.groupby(REGIONS_INDEX).agg(
    {"price": "avg"}
)

# create training data
y_train, y_test = train_test_split(prices_avg)
x_train = embeddings[y_train.index]
x_test = embeddings[y_test.index]

# create regressor from pre-trained hex2vec
regressor = SVR()

# fit regressor to predict rental price from OSM data
regressor.fit(x_train, y_train)

# test results
result = mse(y_test, regressor.predict(x_test))
\end{lstlisting}

\subsection{Transfer learning}

Finally, thanks to the \textit{srai} standardized approach we can easily transfer this model to any other city. We want to find an area with the highest potential for expensive rental in a new city, without rental data. We can do this easily with our model trained in previous steps. We only need to download data for the new region and use a pre-trained model. In this example (presented in Listing \ref{lst:transfer}) we show how to use the pre-trained hex2vec model from the previous step to run predictions on a different city, with the following steps:

\begin{enumerate}
	\item download boundaries for a new city,
	\item load OSM tags based on the filter matching pre-trained model,
	\item preprocess the data following the recipe from Section \ref{sec:examples:processing},
	\item make predictions using the model from the previous step.
\end{enumerate}

Those are simple toy examples, but we hope they show that standardization in the GeoAI domain leads to easier access to advanced modeling tools. Ease of use and unified interfaces mean re-using models becomes almost effortless and available to everyone. We can use models pre-trained at the scale of a whole country/continent without having to create one ourselves. This also may lead to a lower carbon footprint in our experiments. Furthermore, a unified format can be an enabler for benchmarking geospatial models.

\begin{lstlisting}[language=Python, caption=Transfering price regression model to another city, label=lst:transfer]
# get boundaries of new city
area_new = geocode_to_gdf("Hamburg, DE")

# extract OSM filter from pre-trained model
query = embedder.get_query()

# download and process OSM data, same as in stage 1
features_new = ...
regions_new = ...
joint_new = ...
neighbourhood_new = ...

# run predictions
embeddings_new = embedder.transform(
    regions_new,
    features_new,
    joint_new,
    neighbourhood_new,
)

# predict prices
prices_new = regressor.transform(embeddings_new)
\end{lstlisting}

\section{Future works}

This section provides insight into \textit{srai} future releases. We want to show the vision which drives \textit{srai} forward. We intend to extend to other data modalities and focus on making the reproducibility of experiments even more accessible. Below, we present a non-exhaustive list of future functions, which will find their way to the \textit{srai} library.

\subsection{DuckDB with spatial extension as geospatial operations backend}
\label{future:duckdb}

To overcome the memory limitations and speed up the processing of larger datasets, we are currently implementing the most memory-intensive operations using DuckDB with spatial extension. We use it as an in-memory database (which speeds up processing) with secondary file storage (which lifts RAM constraints). This approach has a downside in decreased performance on operations requiring more memory than available. However, it is still a reasonable middle-ground between RAM-limited operations and high-performance clusters because running it locally without extensive configuration is possible.

To keep compatibility with pandas, we implement only the most memory-consuming operations with DuckDB and convert results back to GeoPandas; therefore, switching a backend is transparent from the user's perspective.

\subsection{Pre-trained models and pre-calculated embeddings}

We aim to provide pre-trained models (Hugging Face style\cite{tools:huggingface-models}) and pre-calculated embeddings (word2vec style\cite{word2vec}), which will make geospatial problem-solving accessible for individuals or organizations with limited computation budgets. Another benefit is the opportunity to use transfer learning or benchmark methods using exact embeddings.

It is already possible to load pre-trained models using the \textit{srai} library and use them for any task. However, it is still required to download a model manually. With sufficient resources for hosting and storage, we want to provide hosting for such models and make it possible to download and load them all from the code. We will explore integration with existing solutions (like Hugging Face) or design a specific solution for our library.

Pre-calculated embeddings would allow end-users to solve geo tasks without or with minimal contact with model training. Such embeddings store much information about a region they describe (as shown in hex2vec\cite{our:hex2vec} or loc2vec\cite{related:spatial-embeddings:loc2Vec} with vector addition and subtraction) and may be sufficient to solve some geospatial tasks (e.g., similarity search or unsupervised clustering, as shown in gtfs2vec\cite{our:gtfs2vec}). We intend to create a platform with pre-calculated embeddings for a wide selection of countries and cities, just like pre-calculated word2vec models are available for different languages. 

\subsection{Fine-tuning}

Solving a downstream task is possible by using embeddings to enrich existing data or fine-tuning the whole model for this specific task. In \textit{srai}, we plan to provide an option to add a classification/regression head to any of the available embedders. Those who are trainable could then be fine-tuned for the end task. This way, not only pre-training but also adaptation with fine-tuning will be possible using our library.

\subsection{Other modalities of geospatial data}

Another direction in which \textit{srai} will grow is the support for different data modalities. Right now, we focus on tabular data, mainly from OSM. We already include an option to download map tiles from OSM, so an extension to computer vision (CV) embedding methods is on the roadmap. We will explore integrations with existing libraries like TorchGeo\cite{tools:torchgeo} or implement CV models into \textit{srai}.

The next modality which we want to include is graphs. Geospatial data naturally form graph structures, and some models already rely on graph information for embeddings\cite{related:spatial-embeddings:hrnr, related:spatial-embedings:spatial-networks}. We will include support for graph data download (some are already available with road networks) and dedicated models for graph embeddings designed for geospatial data.

Trajectories are another data modality that can be integrated into srai and benefit from a unified interface. This data type can use embedding methods already existing in the library and link regions that form a trajectory. Trajectories also can be used to build autoregressive models, which is an exciting area of research and would serve as a great addition to the \textit{srai} library.

\subsection{Spatial representations models}

We will continue our work on embedding models for geospatial data and release all of our results compatible with the \textit{srai} library. We hope that more researchers will follow this trend, and just like Hugging Face standardized NLP, the \textit{srai} library will make the first steps to standardize the geospatial AI domain. 

\subsection{Benchmark datasets}

One of the biggest challenges in any domain is the availability of diverse benchmark datasets. They are crucial for experiment reproducibility and comparison between methods. We want to introduce some benchmarks and make them accessible using \textit{srai} library. A good option would be to utilize the Huggingface datasets\cite{tools:huggingface-datasets} with dedicated support for geometries. We hope that this will lay the foundations for standardized geospatial benchmarks.

\subsection{New sources of geospatial data}

We will continue to monitor new emerging platforms for geospatial data. One auspicious example is Overture Maps Foundation\footnote{https://overturemaps.org/}, which provides OSM-like data in PBF format but with finer granularity and more accuracy. Limited data is available as of the time of writing this paper, but if it grows, we will extend our library. Accessing a wide selection of PBF sources under a standard interface will be highly beneficial.

\section{Conclusions}

In this paper, we introduced a new library for geospatial data processing and representation learning - Spatial Representations for Artificial Intelligence (srai). It is a Python module that integrates many geospatial operations and models under a unified API. Its main goal is to unify the process of geospatial data processing and set standards for geospatial model distribution. Hopefully, our library will take the first steps to standardize the geospatial AI domain. 

We extensively reviewed available tools for the geospatial domain and categorized them based on their primary usage. We presented how the \textit{srai} library is positioned against them, which gaps it fills, and how wide the range of tasks it covers. We see that there is a need for this library thanks to its versatility and unified API.

We presented in detail how we designed our library and justified all architectural choices. We described each functionality group and listed available algorithms and models already implemented in the \textit{srai} library. They cover the whole pipeline for most geospatial tasks based on OSM data. We followed that with usage examples demonstrating the capabilities of the \textit{srai} library.

Finally, we shared our view on the future of the library. We intend to expand it in multiple areas, bringing us closer to the primary goal of geospatial AI domain standardization. We believe our work will benefit the whole community and be the foundation for geospatial algorithms and models' unification, reproducibility, and reusability.

\subsection{Limitations}

Right now, due to the costs of computing and hosting, we do not provide pre-calculated embeddings. We hope to solve those issues and provide an option to share and use pre-computed vectors using the \textit{srai} library.

As mentioned in this paper, \textit{srai} currently supports tabular data, which may not always be sufficient for solving all geospatial tasks. However, we made sure that our methods were expandable by using widely adopted data types and libraries. 

The \textit{srai} library relies on GeoPandas\cite{tools:geopandas} for most of the operations, which creates some limitations regarding RAM usage. Currently library must fit the whole spatial dataset into the memory for operations. An area under analysis can be limited by the available resources. There are available multiple solutions that can allow for out-of-memory calculations: dask-geopandas\cite{tools:dask-geopandas} which joins geopandas functionalities with dask capabilities created by the geopandas developers team; Apache Sedona\cite{tools:sedona} library with geospatial operations and the option to connect it to the Apache Spark cluster; DuckDB\cite{tools:duckdb} in-memory database with spatial extension\cite{tools:duckdb_spatial}; and GeoArrow\footnote{\url{https://github.com/geoarrow/geoarrow}} backend for zero-copy operations on geospatial objects. The final option, be it using only one option or implementing multiple backends under a unified API is still to be decided.

\begin{acks}
The free OpenStreetMap data, which is used for the development of \textit{srai}, is licensed under the Open Data Commons Open Database License (ODbL) by the OpenStreetMap Foundation (OSMF).
\end{acks}

\bibliographystyle{ACM-Reference-Format}
\bibliography{references}


\end{document}